\documentclass[lettersize,journal]{IEEEtran}
\IEEEoverridecommandlockouts
\usepackage{cite}
\usepackage{amsmath,amssymb,amsfonts}
\usepackage{algorithm}
\usepackage{graphicx}
\usepackage{textcomp}
\usepackage{xcolor} 
\usepackage{algpseudocode}

\usepackage{multirow}
\usepackage{svg}
\usepackage{subcaption}
\usepackage{graphicx}
\usepackage{makecell}
\usepackage{multirow}
\usepackage{booktabs}
\usepackage{comment}

\usepackage{booktabs}
\usepackage{multirow}
\usepackage{siunitx}
\usepackage{caption}
\usepackage[utf8]{inputenc}

\usepackage{balance}

\usepackage{amsthm}
\def\BibTeX{{\rm B\kern-.05em{\sc i\kern-.025em b}\kern-.08em
		T\kern-.1667em\lower.7ex\hbox{E}\kern-.125emX}}

\begin{document}

\title{
From Edge to Edge: A Flow-Inspired \\ Scheduling Planner for Multi-Robot Systems}

\author{Han Liu, Yu Jin, Mingyue Cui, Boyang Li, Tianjiang Hu, Kai Huang
	
\thanks{
This work was supported by the Guangxi Key R\&D Program (Grant No. GuikeAB24010324), the Guangdong Basic and Applied Basic Research Foundation (Grant No. 2025A1515011485), the Guangdong S\&T Program (Grant No. 2024B1111060004), and the National Natural Science Foundation of China (Grant No. 62232008).
}
\thanks{
		Han Liu, Mingyue Cui, Boyang Li and Kai Huang are with the School of Computer Science, Sun Yat-sen University, Guangzhou, China. (\textit{Kai Huang is  the corresponding author.})}%
	\thanks{ Yu Jin and Tianjiang Hu are with the School of Aeronautics and Astronautics, Sun Yat-sen University, Shenzhen, China.}%
	\thanks{ E-mail: liuh386@mail2.sysu.edu.cn,  huangk36@mail.sysu.edu.cn
	}
}


\maketitle

\hyphenpenalty=4000

\begin{abstract}
Trajectory planning is crucial in multi-robot systems, particularly in environments with numerous obstacles. While extensive research has been conducted in this field, the challenge of coordinating multiple robots to flow collectively from one side of the map to the other—such as in crossing missions through obstacle-rich spaces—has received limited attention.
This paper focuses on this directional traversal scenario by introducing a  real-time scheduling scheme that enables multi-robot systems to move from edge to edge, emulating the smooth and efficient flow of water. Inspired by network flow optimization, our scheme decomposes the environment into a flow-based network structure, enabling the efficient allocation of robots to paths based on real-time congestion levels. The proposed scheduling planner operates on top of existing collision avoidance algorithms, aiming to minimize overall traversal time by balancing detours and waiting times.
Simulation results demonstrate the effectiveness of the proposed scheme in achieving fast and coordinated traversal. Furthermore, real-world flight tests with ten drones validate its practical feasibility. This work contributes a flow-inspired, real-time scheduling planner tailored for directional multi-robot traversal in complex, obstacle-rich environments.

Code: https://github.com/chengji253/FlowPlanner
\end{abstract}

\def\abstractname{Note to Practitioners}
\begin{abstract}
This study aims to address a practical challenge: coordinating multiple robots to traverse from one side of a congested environment to the other—a scenario commonly encountered in applications such as warehouse logistics and urban drone delivery. Traditional motion planning methods often focus on point-to-point navigation and pay limited attention to global flow efficiency under high robot density.
Inspired by flow-based ideas, this work proposes a scheduling planner that structures the environment into a flow-like network and allocates traversal paths based on real-time congestion. 
The method emphasizes directional traversal efficiency, enabling large numbers of robots to move collaboratively and effectively avoid congestion-induced reductions in throughput.
Practitioners deploying multi-robot systems in obstacle-rich environments can integrate this method on top of existing collision avoidance systems to improve overall system efficiency.
\end{abstract}


\section{INTRODUCTION}

Navigating multiple robots from one side of an environment to the other through obstacle-rich spaces is a fundamental yet challenging problem. This scenario is prevalent in a wide range of real-world applications, including warehouse logistics, drone-based delivery systems, and search-and-rescue operations in disaster zones.
In this scenario, efficient trajectory planning for multiple robots is the fundamental issue to be addressed.

Current mainstream approaches typically adopt a two-stage strategy: the front-end generates discrete paths to determine the general motion direction, while the back-end optimizes these paths to achieve collision avoidance. Front-end path planning mainly encompasses three methods:
(a) point-to-point straight-line connections;
(b) single-robot planning (A*);
(c) multi-agent path finding (MAPF) planning.
Point-to-point connections and single-robot planning overlook inter-robot interactions, delegating collision avoidance entirely to the back-end. In contrast, MAPF accounts for inter-robot coordination but is constrained by grid-based assumptions, resulting in paths with insufficient flexibility.
Recent research has primarily focused on back-end optimization, addressing collision avoidance, reducing computational load, and minimizing deadlocks.
Nevertheless, we find that there is still much room for improvement in the front-end path planning, especially in the case where multiple robots need to rapidly traverse an environment with obstacles.

Achieving rapid passage of multi-robot systems through obstacle-rich areas requires maximizing the utilization of every available path. This concept is similar to network flow problems, which aim to maximize flow from source to sink in a network \cite{ford1958constructing}. The core of network flow lies in optimizing network capacity utilization to enhance flow and speed. 
We observe that the idea from maximum flow problems can be applied to the front-end discrete path planning, accelerating their traversal through obstacle-rich regions.

However, applying flow-based methods to multi-robot systems still faces numerous challenges. 
The reasons are as follows: 
First, flow-based methods require a predefined network structure, but real-world map environments often lack such readily usable network structures.
Second,  these methods typically have long computation times and are often designed for offline use, lacking real-time applicability.
Furthermore, flow-based methods typically produce flow distributions over edges rather than directly executable trajectories for individual robots. 
This limitation hinders their integration into the prevalent two-stage architecture of robotic systems.
Therefore, there is an urgent need for a front-end planner that not only provides an executable path for each robot, but also leverages flow-based principles to improve efficiency, while maintaining compatibility with back-end collision avoidance modules.

To address these challenges, we propose a flow-based multi-robot planning scheme.
The scheduling planner, as a front-end module, provides discrete paths for each robot in real-time, guiding their direction of movement, and is compatible with back-end collision avoidance modules.
Our scheme first decomposes the initial map and extracts information to form a network containing nodes and edge information.
Subsequently, the planner calculates the path for each robot based on the current position of the robots and the congestion level of the map channels. 
The planner optimizes path selection from the perspective of the entire swarm, balancing detouring and waiting to minimize the time required for robots to traverse obstacle-rich environments. Experimental results show that our planner has high computational efficiency.
And we conducted flight tests to demonstrate that the proposed scheme can be applied to real-world settings.

Our contributions can be summarized as follows:
\begin{itemize}
	\item We propose a scheduling planner that utilizes the traversable areas of the map, thereby reducing the traversal time for multiple robots in obstacle-rich environments.
	
	\item The proposed scheduling planner achieves high computational efficiency and supports real-time online operation as a front-end module for multi-robot navigation.
	
	\item We validate the effectiveness and practicality of the proposed framework through comprehensive simulations and real-world drone experiments, demonstrating improved traversal efficiency compared to representative state-of-the-art methods.

\end{itemize}

The rest of this paper is organized as follows. We introduce the related work in Section II.
In Section III, we introduce the statement of the problem and the framework of our scheme.
Then, we present the details of network construction in Section IV.
The details of the flow planner are discussed in Section V.
In the following, the results of the experiments are presented in Sections VI. 
Finally, we draw conclusions and discuss our future work in Section VII.

\section{related work}

In multi-robot trajectory planning research, the mainstream framework typically divides the process into a front-end and a back-end stage.
The front-end generates a discrete path for each robot, determining the general direction of motion.
The back-end then refines these paths into smooth trajectories that satisfy dynamic constraints and ensure collision avoidance via optimization.
Over the past few years, several representative back-end planning methods have been developed, including artificial potential field methods~\cite{6425500,9108576}, velocity obstacle (VO) approaches and their variants~\cite{RVO,HRVO}, and optimization-based frameworks such as on-demand collision avoidance~\cite{8950150,9199826} and gradient-based local planning~\cite{9561902}.
Distributed model predictive control has also received increasing attention~\cite{soria2021predictive,wang2014synthesis}.
More recent efforts have focused on improving real-time performance and robustness.
For example, Tordesillas \textit{et al.}~\cite{2021mader} propose an asynchronous planner that leverages inter-agent communication to ensure the generation of collision-free trajectories.
Park \textit{et al.}~\cite{park2023dlsc} present a deadlock-free trajectory planning algorithm for quadrotor swarms operating in maze-like environments.
Charbel \textit{et al.}~\cite{toumieh2024high} introduce a high-speed, decentralized motion planning framework for aerial swarms, ensuring safe navigation in unknown environments.
These works primarily target the back-end, aiming to improve efficiency in collision avoidance, prevent deadlocks, and handle complex scenarios.

On the other hand, their front-end are generally kept simple and falls into one of the following three categories:
(a) Point-to-Point: a straight-line connection from start to goal;
(b) Single-robot planning: standard planners such as A*, jump point search are applied independently to each robot;
(c) Multi-Agent Path Finding (MAPF): discrete paths are computed using multi-agent coordination strategies.
For instance, Enhanced Conflict-Based Search (ECBS) is adopted in~\cite{9197162,9293348} for front-end planning.
Point-to-Point and single-robot planning largely ignore inter-robot interactions, leaving all collision avoidance to the back-end.
In contrast, MAPF-based approaches perform time-space coordination in the front-end, thus reducing conflicts.
However, MAPF has notable limitations.
It assumes grid-based, one-step-at-a-time movements, leading to rigid paths that perform poorly when integrated with the back-end.
Moreover, all three front-end methods are typically executed offline, thus lack adaptability to real-time disturbances such as congestion.

Research on the MAPF problem spans decades, resulting in numerous classic algorithms. 
Among them, Conflict-Based Search (CBS) and its variant (ECBS) \cite{ECBS} are a class of widely popular optimal algorithms that first plan paths for each agent independently and then resolve collisions through high-level search.
P-SIPP \cite{p-sipp} is another classic algorithm that combines safe interval path planning \cite{sipp} with a priority mechanism to plan paths.
However, with classical algorithms, as the number of robots increases or the scenario becomes more complex, the success rate becomes very low.
Recent MAPF advancements have improved scalability and solution quality for large-scale instances. Among optimal and bounded-suboptimal algorithms, EECBS \cite{EECBS} and Lazy CBS \cite{LazyCBS} enhance speed by optimizing conflict detection and constraint generation, while CBSH2-RTC \cite{CBSH2-RTC} further improves CBS efficiency with heuristic functions and runtime conflict checking.
LNS \cite{LNS1} improves MAPF solutions via large neighborhood search, replanning paths for agent subsets. LNS2 \cite{LNS2} enhances this by generating and repairing infeasible plans, demonstrating flexibility and efficiency in complex settings.

In recent years, several studies have addressed congestion-aware multi-robot planning. Charlie \textit{et al.}~\cite{TVMA} proposed a strategy synthesis framework that models each robot with a time-varying markov automaton and solves for policies using heuristic dynamic programming. This approach effectively captures dynamic congestion along paths, but it operates offline, has high computational overhead, and scales only to small teams ($ \leqslant15$ robots). CMPP \cite{CMPP} formulates the problem as congestion mitigation on sparse network graphs and employs a two-layer branch-and-bound–like search for efficient planning. While effective at reducing conflicts and delays in large-scale settings, CMPP requires manually predefined sparse graphs and its cost function considers congestion alone, often leading to excessive detours and increased travel distance.

\begin{figure*}[t]
	\centering
	\includegraphics[scale=0.90]{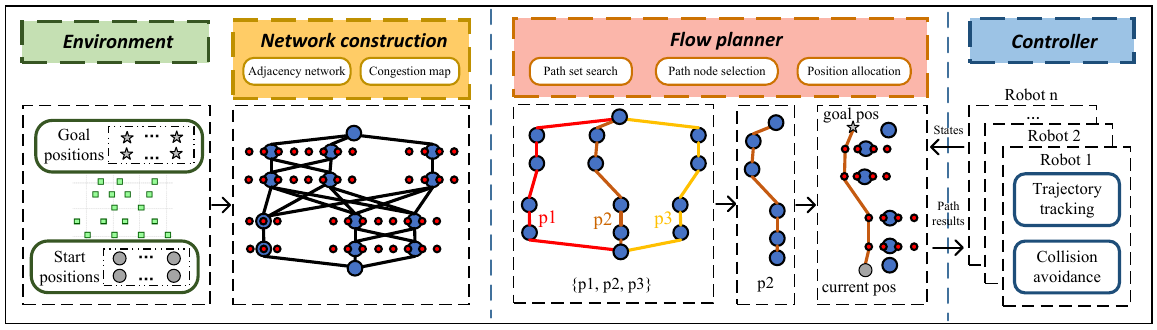}
	\caption{The framework of our scheme.}
	\label{fig:algo}
\end{figure*}

Since the seminal work of Ford and Fulkerson \cite{ford1958constructing}, network flow problems have been a cornerstone of operations research. With expanding applications, many variants have emerged, such as quickest flow \cite{quickest-flow}, minimum-cost flow \cite{Minimum-cost-flow}, time-dependent flow \cite{sung2000shortest}, and multi-commodity network flow (MCNF) \cite{bevrani2020multi}.
The MCNF problem aims to route multiple commodities through a capacity-constrained network at minimal cost. It parallels multi-robot scheduling, where multiple agents compete for shared resources under capacity constraints, typically minimizing total cost or makespan.
Numerous methods have been proposed for MCNF, including benders decomposition \cite{mahey2001capacity}, column generation \cite{trivella2021multi}, approximation algorithms \cite{oldham2001combinatorial}, and meta-heuristics \cite{gen2008network}. 
However, these methods face challenges when applied to robotic systems: benders decomposition and column generation methods are computationally intensive, meta-heuristics may suffer from slow convergence in large robotic scenarios, and approximation algorithms focus more on theoretical bounds than practical applicability.
Moreover, classical flow models abstract movement as units of flow, overlooking the kinematic and collision-avoidance constraints in robotics.
Although limited in number, there have been some attempts to incorporate flow-based ideas into robotic planning. 
Yu \textit{et al.} \cite{yu2013multi,yu2013planning} offer an elegant reduction of MAPF to a flow-based model, enabling the use of optimization solvers for optimality. 
However, as their method operates offline and is limited in scalability ($ \leqslant50$ robots), it struggles in large-scale contexts.
Dugas \textit{et al.}\cite{dugas2022flowbot} modeled crowds as a flow model and proposed a method for robot navigation within crowds. Janchiv \textit{et al.}\cite{janchiv2013time} applied flow network techniques to the multi-robot complete coverage path planning problem. 
Despite these efforts, no existing method has successfully leveraged flow-based ideas for real-time planning in multi-robot systems.

These observations motivate our proposed flow-inspired scheduling planner, which operates at the front-end to provide real-time, congestion-aware discrete path  for robots navigating from one side of an obstacle-rich environment to the other, enabling smooth integration with back-end collision avoidance modules.

\section{Problem Statement}

We consider a two-dimensional workspace \(\mathcal{W} \subseteq \mathbb{R}^2\) populated with a set of \(M\) static obstacles \(\mathcal{O} = \{o_1, o_2, \ldots, o_M\}\), where each obstacle \(o_i \subset \mathcal{W}\). A team of \(N\) robots, indexed by \(n \in \mathcal{N} := \{1, 2, \ldots, N\}\), must traverse the environment from their respective start locations \(\mathbf{s}^{(n)} \in \mathcal{W}\) to goal locations \(\mathbf{g}^{(n)} \in \mathcal{W}\).
In this work, we focus on a scenario where all start positions are located on one side of the environment, and all goal positions are located on the opposite side. This models a broad class of practical applications such as coordinated street crossing, indoor-to-outdoor transitions.
Each robot is modeled as a point mass in \(\mathbb{R}^2\) with single integrator dynamics. Let \(\mathbf{p}^{(n)}(t)\) and \(\mathbf{v}^{(n)}(t)\) denote the position and velocity of robot \(n\) at time step \(t\), respectively. Given a fixed time discretization step \(h > 0\), the motion model is described by:
\begin{eqnarray}
\mathbf{p}^{(n)}(t+h) = \mathbf{p}^{(n)}(t) + h \mathbf{v}^{(n)}(t).
\end{eqnarray}
The velocity of each robot is bounded due to physical actuation limits: $\|\mathbf{v}^{(n)}(t)\| \leq \mathbf{v}_{\max}$.
To ensure safety during navigation, robots must maintain a minimum distance \(r_{\min}\) from one another and from all obstacles:
\begin{align}
\left\|\mathbf{p}^{(i)}(t)-\mathbf{p}^{(j)}(t)\right\| & \geq r_{\min }, i, j \in \mathcal{N}, i \neq j, \\
\left\|\mathbf{p}^{(i)}(t)-o_j\right\| & \geq r_{\min }, i \in \mathcal{N}, o_j \in \mathcal{O} .
\end{align}
The objective of the system is to coordinate all robots to reach their respective goals as efficiently as possible. The overall mission completion time is defined as the time at which the last robot arrives at its destination. 
This min-max objective captures the makespan of the multi-robot traversal and serves as a measure of collective efficiency.
Thus, the core problem is to develop a scheduling and planning strategy that allocates robots to different paths, considering the path capacities, in such a way that the time for the last robot to reach the goal is minimized. This requires a sophisticated scheme to ensure that the system operates efficiently in the presence of complex constraints.

\subsection{Overview}

Before delving into the technical details, we briefly outline the overall structure of our scheme, as illustrated in Fig.   \ref{fig:algo}. 
The leftmost part represents the Environment, which models the operational space, including static obstacles and navigable areas.
Adjacent to it is the network construction module, which provides two core map representations: the adjacency network and the congestion map. These are used to model feasible robot paths and regional capacity constraints, serving as a bridge between environmental information and the planner.
At the core lies the flow planner, which comprises three sequential components:
path set search,
path node selection,
and position allocation.
These modules work in tandem to generate suitable trajectories for each robot.
Finally, each robot is equipped with a controller, consisting of two modules: trajectory tracking and collision avoidance. Once a planned path is received, the controller executes the path while ensuring safety through collision avoidance, which takes priority over trajectory tracking.
For multi-robot collision avoidance, we employ the ORCA \cite{RVO} algorithm — a robust, velocity obstacle method that remains effective even with a large number of agents.

\section{Network construction}

This section provides a detailed description of the network construction method. In robot navigation, grid maps are commonly used to represent the environment, whereas flow-based algorithms typically rely on network structures composed of nodes and edges, revealing a fundamental structural difference between the two representations.
Therefore, it is essential to establish a bridge between flow-based algorithms and grid maps. The key to building this bridge lies in extracting core information from the map, including paths, capacities, connectivity, and path lengths. Proper abstraction and refinement of the map data form the foundation for efficient planning.

\subsection{Adjacency network and congestion map}

To enable flow-based robot scheduling and planning, we propose two key map representations: the adjacency network and the congestion map.
The adjacency network is a network model that characterizes  potential movement paths and their passage capacities through a structure of nodes and edges.
The congestion map partitions the spatial environment into multiple sub-regions to analyze the distribution of robots and capacity constraints within each region.

\begin{figure}[t]
	\centering
	\includegraphics[scale=0.43]{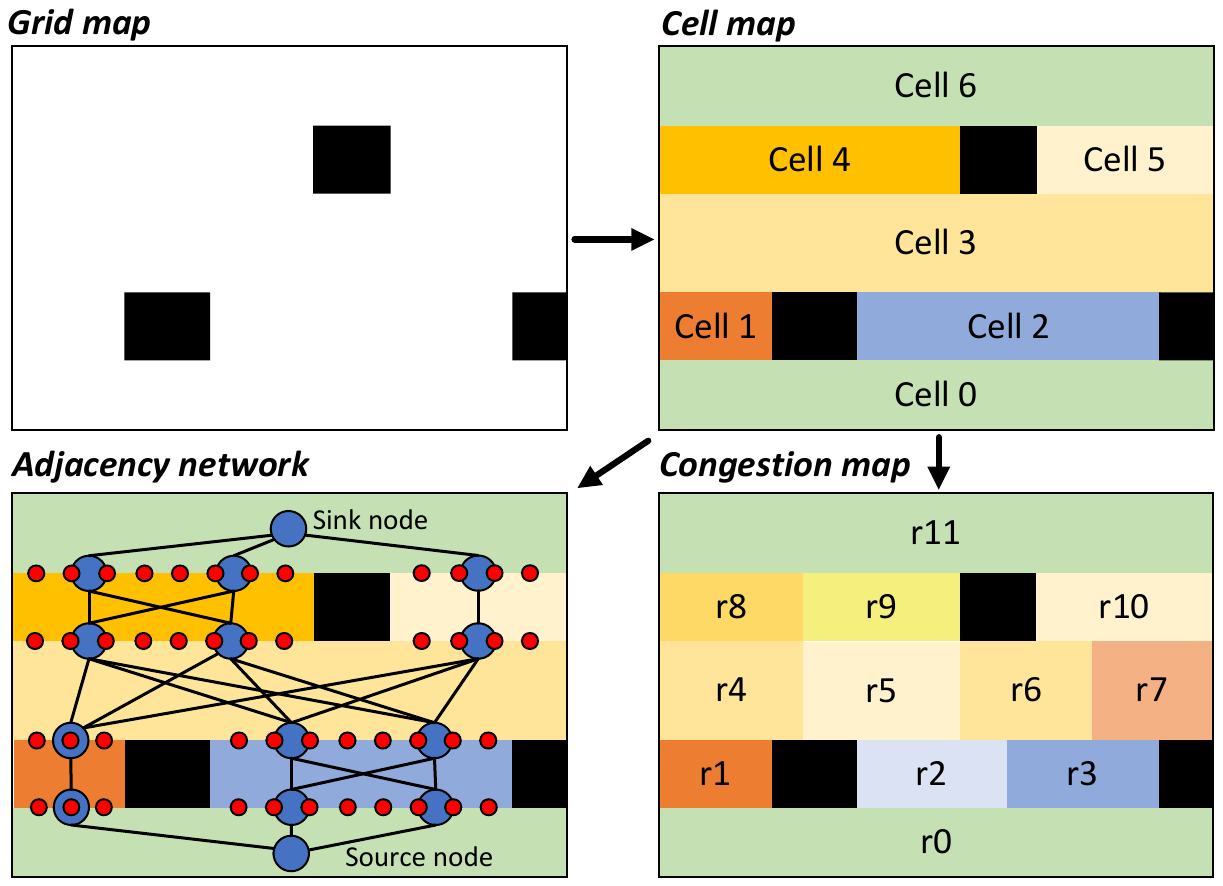}
	\caption{The network construction process. In the adjacency network, the blue circles represent the $\operatorname{Node}$ of the network, and the red circles represent the $\operatorname{Pos}$ of each node.}
	\label{fig:mapinfo}
\end{figure}

The adjacency network models potential robot movement paths using a graph structure \(\mathcal{G}=\{\mathcal{V}, \mathcal{L}\}\), where \(\mathcal{V}\)
is the set of nodes and \(\mathcal{L}\)
 is the set of edges.
The construction process begins with the Boustrophedon Cellular Decomposition method \cite{choset2000coverage}. 
This method scans the map back and forth, akin to an ox plowing a field, dividing the map into multiple cells. The collection of all cells is denoted as \( C \).
Each cell contains local connectivity information and boundary details.
The resulting decomposed map is defined as the cell map.
Next, we build the adjacency network from the cell map. Specifically, we identify adjacency boundaries between neighboring cells and uniformly sample these boundaries with a number of key positions.
The number of positions generated on the boundary is
$\frac{L_B}{\phi r_{\min }}$,
where ${L_B}$ is the boundary length, $r_{\min }$ is the minimum robot safety radius, and 
$\phi \in[1,2]$ 
is a redundancy factor. 
The resulting key positions are termed \(\operatorname{Pos}\), and intuitively represent the capacity of the boundary. 
Then, we group every $N_B$
consecutive 
\(\operatorname{Pos}\) to form a \(\operatorname{Node}\),
where $N_B$ denotes the number of positions aggregated to form one node.
The coordinate of each \(\operatorname{Node}\)  is defined as the mean position of its constituent \(\operatorname{Pos}\).
These \(\operatorname{Node}\)s become the nodes in the adjacency network. Then, we create edges between nodes based on their relative positions: if two \(\operatorname{Node}\)s belong to the upper and lower boundaries of the same cell respectively, an edge connection is established between them.
To clearly distinguish and facilitate identification, in the subsequent sections of this paper, when using \(\operatorname{Node}\) and \(\operatorname{Pos}\), we specifically refer to the nodes and positions in the adjacency network, represented in the Fig.  
\ref{fig:mapinfo} by blue and red circles, respectively.

While the cell map captures connectivity, it may not provide fine-grained resolution for congestion analysis—especially when cell sizes vary significantly. To address this, we further divide each cell into uniform sub-regions based on a preset unit length \(L_{con}\)
and width \(W_{con}\), unless the cell already meets the desired spatial granularity.
This results in a congestion map, denoted by the region set \( \mathcal{R} = \{1,2, \ldots, R\} \), where each region $r \in \mathcal{R}$ has an area $\operatorname{Are}(r)$ and current robot count $\operatorname{Num}(r)$ . The capacity of each region is estimated by,
$\operatorname{Cap}(r)=\frac{\operatorname{Are}(r)}{\left(\phi r_{\min }\right)^2}$.
Due to the redundancy factor $\phi$, this capacity underestimates the theoretical maximum to leave space for dynamic movement and avoid congestion-induced performance degradation.
Fig. \ref{fig:mapinfo} shows a network construction example.
In the figure, we illustrate how a simple grid map is processed through map decomposition and network construction to obtain the defined adjacency network and congestion map.

The reason for using two separate abstractions—the adjacency network and the congestion map—instead of building a single unified network with nodes, edges, and capacities, is to improve modeling accuracy. In the adjacency network, some edges may overlap or intersect spatially, occupying the same physical area. Assigning capacities directly to each edge could lead to double-counting the same space, resulting in inaccurate capacity estimates. Therefore, we separate path connectivity and area capacity into two representations, enabling a more precise modeling of shared spatial constraints.

\subsection{Network-flow formulation}

Given a network $\mathcal{G}=\{\mathcal{V},\mathcal{L}\}$, we formulate multi-robot path planning as a network-flow problem. 
Each robot $n$ is assigned a start–goal pair $(s^{(n)}, g^{(n)}), s^{(n)},g^{(n)}\in\mathcal{V}
$.
The path of robot $n$ is represented as a vertex sequence:
$
\mathrm{Node}^{(n)}=\{v[1],v[2],\ldots,v[Q]\},\quad v[1]=s^{(n)},\; v[Q]=g^{(n)}
$, where $Q$ is the number of nodes.
And the set of all robot paths is
$
\pi=\{\mathrm{Node}^{(1)},\mathrm{Node}^{(2)},\ldots,\mathrm{Node}^{(N)}\}.
$
We introduce binary decision variables
$
z_{l}^{(n)}\in\{0,1\},\quad n\in\mathcal{N},\; l\in\mathcal{L},
$
indicating whether robot $n$ traverses edge $l$. Let $\delta^{-}(v)$ and $\delta^{+}(v)$ denote the sets of incoming and outgoing directed edges of node $v$. 
These variables are subject to several constraints. 
First, each robot must leave its start node and enter its goal node exactly once:
\begin{eqnarray}
\sum_{l\in\delta^+(s^{(n)})}z_{l}^{(n)}=1,\quad
\sum_{l\in\delta^-(g^{(n)})}z_{l}^{(n)}=1,\qquad
\forall n\in\mathcal{N}.
\label{eq:44}
\end{eqnarray}
To ensure path continuity, flow conservation is enforced at every intermediate node:
\begin{eqnarray}
\sum_{l\in\delta^-(v)}z_{l}^{(n)}=\sum_{l\in\delta^+(v)}z_{l}^{(n)},
\quad \forall v\in\mathcal{V}\setminus\{s^{(n)},g^{(n)}\}.
\label{eq:45}
\end{eqnarray}
For each undirected physical edge, a robot may choose at most one travel direction, preventing self-conflicts (simultaneous use of both directions by the same robot):
\begin{eqnarray}
z_{(u,v)}^{(n)}+z_{(v,u)}^{(n)}\leq 1,\quad
\forall n\in\mathcal{N},\;\forall (u,v)\in\mathcal{L}.
\label{eq:46}
\end{eqnarray}
Let $T^{(n)}$ denote the total time required for robot $n$ to travel to its goal. This arrival time depends not only on the chosen path but also on dynamic interactions such as local collision avoidance and congestion. Formally,
\begin{eqnarray}
T^{(n)}=\mathcal{F}(\pi,\mathcal{W},\mathcal{A}),
\label{eq:47}
\end{eqnarray}
where $\mathcal{W}$ denotes workspace/map information, $\mathcal{A}$ the collision-avoidance module, and $\pi$ the set of all robot paths. The functional $\mathcal{F}$ aggregates path length, local avoidance effects, regional capacity, and congestion induced by simultaneous occupancy with other robots. Although $\mathcal{F}$ is not available in closed form, it plays a central role in the optimization.
The objective is to minimize the makespan, i.e., the maximum arrival time among all robots:
$$
\mathrm{minimize}\;\; \max_{n\in\mathcal{N}} T^{(n)},
$$
subject to \eqref{eq:44}, \eqref{eq:45},
\eqref{eq:46}, \eqref{eq:47}.

The formulation bears similarity to classical network flow problems but departs in key aspects.
(i) The edge-selection variables $z_{l}^{(n)}$ instantiate the routing of unit commodities (robots). (ii) The start/goal constraints and flow conservation encode the standard source/sink balance and intermediate-node balance, respectively. (iii) The unique-direction constraint on undirected edges eliminates intra-robot self-conflicts. (iv) Capacity budgeting induced by the congestion map acts as a shared-resource constraint, avoiding double-counting capacity of the same physical space. Unlike static network-flow models, our objective is implicitly coupled to dynamic congestion and interaction effects through $\mathcal{F}$ and the avoidance module $\mathcal{A}$, yielding a cost that more faithfully reflects execution-time behavior.

It should be noted that since the network only provides a sparsified abstraction of the workspace, in the subsequent method section each $\mathrm{Node}^{(n)}$ is further refined into a high-resolution position sequence $\mathrm{Pos}^{(n)}$ to enhance execution quality. 
In addition, we do not explicitly specify motion timing; paths are represented solely as ordered spatial sequences, with timing emerging during execution from robot-level control and local collision avoidance.

\section{Flow Planner}

In this section, we present the detailed design of our flow planner. 
The flow planner is composed of three key components: path set search, path node selection, and  position allocation.
The path set search module identifies a set of potential candidate paths based on the current environment and task configuration.
The path node selection module evaluates these candidate paths, balancing path length and expected waiting time to select the most suitable trajectory.
Finally, the position allocation module determines the precise entry and exit positions on the selected path, ensuring smooth transitions and minimal conflict.
These modules are executed sequentially, producing a valid and congestion-aware path for each robot, which is then transmitted to the controller for execution.

\subsection{Path set search}

To enable efficient path  selection for each robot, we propose a path set search method that constructs a subset of viable paths from which the optimal trajectory can later be selected. This step is critical because, as the complexity of the map and the number of obstacles increase, the total number of feasible paths in the network grows exponentially with the size of the environment. Exhaustively enumerating all possible paths is computationally prohibitive and undermines the real-time requirements of multi-robot systems. Hence, it is essential to identify a promising subset of candidate paths that maintains decision quality while significantly reducing computational overhead.

Let the next-go-to path set of the 
$n$-th robot at time $t$ 
be denoted as $\mathcal{P}^{(n)}_{\text{next}}(t)$.
The following procedure outlines our method for constructing this path set (Algorithm \ref{alg:path_set_search}):
(a)  Identify all upper boundary nodes of the cell currently occupied by robot $n$, denoted by
$\operatorname{UP}^{(n)}$ (line 2).
(b) Calculate the distances between the current position of the robot and these boundary nodes, and divide them into two groups according to the distances: the $\alpha$ closest nodes form the set $\operatorname{UP}^{(n)}_{\text{clos}}$, and the remaining nodes are represented as the set $\operatorname{UP}^{(n)}_{\text{left}}$ (line 3).
(c) Find all the lower boundary nodes $\operatorname{DN}^{(n)}$ of the cell where the target position of the robot is located (line 4).
(d) Calculate the distances between the target position and all the nodes in $\operatorname{DN}^{(n)}$, and the $\beta$ closest nodes among them are represented as the set $\operatorname{DN}^{(n)}_{\text{clos}}$ (line 5).
(e) Use Dijkstra's algorithm to find all the paths with all the nodes in the set $\operatorname{UP}^{(n)}_{\text{clos}}$ as the starting points and all the nodes in the set $\operatorname{DN}^{(n)}$ as the target nodes (line 7).
(f) Similarly, compute all shortest paths from  $\operatorname{UP}^{(n)}_{\text{left}}$ to $\operatorname{DN}^{(n)}_{\text{clos}}$ (line 8).
By combining the results of these two path construction steps, we obtain the final candidate path $\mathcal{P}^{(n)}_{\text{next}}(t)$.

\begin{algorithm}[t]
	\caption{Path set search}
	\label{alg:path_set_search}
	\begin{algorithmic}[1]
		\State \textbf{Input:} Current position \(\mathbf{p}^{(n)}(t)\), target position $\mathbf{g}^{(n)}$
		\State Identify upper boundary nodes $\operatorname{UP}^{(n)}$
		\State Identify $\operatorname{UP}^{(n)}_{\text{clos}}$ and $\operatorname{UP}^{(n)}_{\text{left}}$ based on distance to $\mathbf{p}^{(n)}(t)$
		\State Identify lower boundary nodes $\operatorname{DN}^{(n)}$
		\State Identify $\operatorname{DN}^{(n)}_{\text{clos}}$ based on distance to $\mathbf{g}^{(n)}$
		\State Use Dijkstra algorithm to find all paths from:
		\State \quad (a)  $\operatorname{UP}^{(n)}_{\text{clos}}$ to $\operatorname{DN}^{(n)}$
		\State \quad (b) $\operatorname{UP}^{(n)}_{\text{left}}$ to $\operatorname{DN}^{(n)}_{\text{clos}}$
		\State \textbf{Output:}  $\mathcal{P}^{(n)}_{\text{next}}(t)$
	\end{algorithmic}
\end{algorithm}

The algorithm starts from the robot’s current cell and directly plans paths from exit nodes (upper boundary) to entrance nodes (lower boundary) of the target cell, avoiding redundant exploration of intermediate branches. By prioritizing exits near the robot and entrances close to the destination, and incorporating these connections into the search set, it reduces computational complexity and improves global planning efficiency.
To meet real-time requirements, the algorithm does not exhaustively explore all possible paths but instead focuses on representative and high-probability ones. It addresses two key aspects: (1) exiting the current cell efficiently, and (2) entering the target cell quickly. Following a “start-to-exit, entrance-to-destination” strategy, it minimizes attention to intermediate details and enhances overall practicality.
A distance-based filtering mechanism further improves the quality of the path set. Exits closer to the robot yield shorter initial paths, while entrances near the target allow faster convergence. This reduces computational overhead while retaining promising candidate paths, ensuring real-time performance and providing a representative, high-quality path set for subsequent decision-making.

\subsection{Path node selection}

After obtaining the set of candidate paths $\mathcal{P}^{(n)}_{\text{next}}(t)$, the next step is to select one path to assign to $\operatorname{Node}^{(n)}$ as the control variable for the $n$-th robot. 
In this section, we present a path selection strategy based on mixed-integer programming.

\begin{figure}[t]
	\centering
	\includegraphics[scale=1.5]{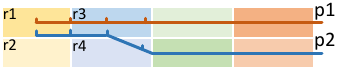}
	\caption{An example of $K$ sub-segments. In this example, $K = 3$. The initial segment of each path is extracted and divided into three equal parts. For path $p_1$, the three sub-segments correspond to regions $r_1, r_3,$ and $r_3$, respectively. For path $p_2$, the regions are $r_1, r_3,$ and $r_4$.
	This suggests that if two robots follow paths $p_1$ and $p_2$, there is a high likelihood of spatial overlap in regions $r_1$ and $r_3$, potentially leading to increased congestion in those areas.}
	\label{fig:predict-K}
\end{figure}

Let $p$ denote a candidate path element in the set $\mathcal{P}^{(n)}_{\text{next}}(t)$. We introduce a binary decision variable $z_{p}^{(n)}(t)$, which indicates whether the $n$-th robot selects path $p$ (i.e., $z_{p}^{(n)}(t) = 1$ if selected, and $0$ otherwise). The set of all decision variables is defined as:
\begin{eqnarray}
	\mathcal{Z} = \left \{ z_{p}^{(n)}(t)  \mid n \in \mathcal{N},  p \in \mathcal{P}^{(n)}_{\text{next}}(t) \right \}
\end{eqnarray}
where $\mathcal{N}$ represents the set of all robots.
Since each robot must select exactly one path at any given time, we impose the following constraint:
\vspace{-0.2cm}
\begin{eqnarray}
	\sum_{p \in \mathcal{P}^{(n)}_{\text{next}}(t) }^{}  z_{p}^{(n)}(t)  = 1.
	\label{eq:cons 1}
\end{eqnarray}
To model the collective path selection behavior across all robots, we define the union of all path sets as:
\begin{eqnarray}
	\mathcal{P}^{}_{\text{next}}(t) =\bigcup\left\{\mathcal{P}^{(n)}_{\text{next}}(t) , n \in \mathcal{N} \right\}.
	\label{eq:next path set all}
\end{eqnarray}
To predict potential congestion among different candidate paths, we extract the initial segment of each path $p \in \mathcal{P}_{\text{next}}(t)$ with a fixed spatial length $L_{\text{pre}}$. This segment is uniformly divided into $K$ sub-segments, each of length $\frac{L_{\text{pre}}}{K}$. Since a sub-segment may traverse multiple regions on the congestion map, we assign a representative region to each sub-segment by selecting the region in which the path occupies the longest portion. The resulting region sequence for path $p$ is denoted as $\{r_1, r_2, \dots, r_k, \dots, r_K\}$, where $r_k$ represents the dominant region of the $k$-th sub-segment. This region sequence is then used for subsequent analysis of potential conflicts and congestion among the candidate paths.
Fig. \ref{fig:predict-K} shows an easy example of $K$ sub-segments.

By extracting the front part of each path $p$ in this way, we aim to analyze the potential future regions the robot may reach.
If the robot selects path $p$, then region $r_k$ represents the region it may reach in the $k$-th time interval. If multiple robots arrive at the same region simultaneously, congestion may occur.
We define the function $\operatorname{Occ}(p, k) = r$ to indicate that the $k$-th segment of path $p$ occupies region $r$. 
The set of regions occupied by all paths in $\mathcal{P}_{\text{next}}(t)$ during the $k$-th sub-segment is:
\begin{eqnarray}
\mathcal{R}_k(t)=\left\{\operatorname{Occ}(p,k) \mid p \in \mathcal{P}_{\text {next }}(t)\right\} .
\end{eqnarray}
Additionally, we define the subset of paths in $\mathcal{P}_{\text{next}}(t)$ that occupy region $r$ in the $k$-th sub-segment as:
\begin{eqnarray}
\mathcal{P}_{\text {k }}^r(t)=\left\{p \mid p \in \mathcal{P}_{\text {next }}(t), \operatorname{Occ}(p, k)=r\right\} .
\end{eqnarray}
With these definitions, we now quantify the impact of path selection on regional congestion. 
We introduce a concept called the regional traffic load $L_k^r(t)$, which is defined as:
\begin{eqnarray}
\max(\sum_{n \in \mathcal{N}} \sum_{p \in \mathcal{P}_k^r(t)} z_p^{(n)}(t)+\operatorname{Num}(r)-\operatorname{Cap}(r), 0),
\end{eqnarray}
where $\sum_{n \in \mathcal{N}} \sum_{p \in \mathcal{P}_k^r(t)} z_p^{(n)}(t)$ calculates the number of paths chosen by all robots to pass through region $r$, $\operatorname{Num}(r)$ is the current occupancy of the region, and $\operatorname{Cap}(r)$ is the capacity of this region.
The $\max(\cdot, 0)$ function ensures that only overload conditions are penalized, not unused capacity.
$L_k^r(t)$ reflects the degree of load of region $r$ in the $k$-th sub-segment.
To measure the overall congestion level, we define the queuing cost:
\begin{eqnarray}
f_{\text {Que }}(t)=\sum_{k=1}^K \omega_k \sum_{r \in \mathcal{R}_k(t)} \frac{\left[L_k^r(t)\right]^2}{C_r} .
\end{eqnarray}
where $C_r = \operatorname{Cap}(r)^2$ is the normalization factor, and $\omega_k$ is the weight coefficient of the $k$-th segment, which is used to emphasize the contribution of different segments to the congestion cost. Squaring the overload amount is to amplify the cost when the region is overloaded, so as to penalize the paths that lead to congestion and encourage the path allocation to avoid overload situations as much as possible. 
If we want to pay more attention to the congestion situation close to the current moment, we can set $\omega_k$ to decrease with $k$. If all segments are equally important, we can set the weights to be all 1.

In addition to congestion, we account for the path length cost, defined as:
\begin{eqnarray}
f_{\operatorname{run}}(t) = \omega_{\operatorname{run}} \sum_{n\in \mathcal{N}} \sum_{p \in \mathcal{P}^{(n)}_{\text{next}}(t)} \operatorname{Len}(p) z_p^{(n)}(t)
\end{eqnarray}
where $\operatorname{Len}(p)$ represents the length of path $p$, and $\omega_{\operatorname{run}}$ represents the weight coefficient of the path length cost.
Finally, we combine all the cost terms to obtain the final optimization objective:
\begin{eqnarray}
\underset{\boldsymbol z_p^{(n)}(t)}{\operatorname{minimize}} \quad f_{\operatorname{Que}}(t) + f_{\operatorname{run}}(t)
\end{eqnarray}
By solving this optimization problem, we obtain the optimal path selection for each robot, ensuring efficient path allocation and minimizing the likelihood of regional overload by balancing congestion and path length considerations.

\subsection{Position allocation}

\begin{figure}[t]
	\centering
	\includegraphics[scale=1.1]{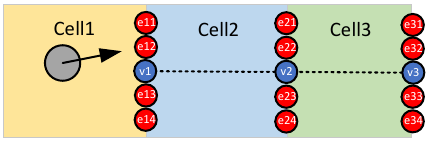}
	\caption{An example of position allocation.
	After obtaining the result \(\operatorname{Node}^{(n)} = \{v_1, v_2, v_3\}\) through the path node selection process, each node offers $N_B=4$ candidate \(\operatorname{Pos}\) options. Thus, there are a total of \(4^3\) possible combinations. The position allocation selects the final set from these candidates, for example, \(\operatorname{Pos}^{(n)} = \{e_{12}, e_{23}, e_{34}\}\), which defines the specific path the robot should follow at time \(t\).}
	\label{fig:position example}
\end{figure}

Following the path node selection phase, we determine the path node sequence \(\operatorname{Node}^{(n)} = \{v[1], v[2], \dots, v[Q]\}\) for each robot \(n\), where \(v[i]\) denotes the \(i\)-th node, $Q$ denotes the number of nodes.
As each node \(v[i]\) corresponds to a set of optional positions \(E_{v[i]}\), we aim to assign a specific position sequence \(\operatorname{Pos}^{(n)} = \{e[1], e[2], \dots, e[Q]\}\), where \(e[i] \in E_{v[i]}\), to each robot \(n\), thereby generating the final discrete path. 
For notational convenience, we denote the $i$-th node is $\operatorname{Node}^{(n)}$ as \(\operatorname{Node}^{(n)}[i]\).
Fig. \ref{fig:position example} shows a simple example of position allocation.
With each node providing at least \(N_B\) optional positions, efficiently and equitably selecting these positions is an essential step for ensuring balanced resource allocation.
To address this challenge, we propose a heuristic greedy algorithm that jointly considers distance cost and position occupancy to generate an effective position allocation.
The details of our algorithm are as follows (Algorithm~\ref{alg:path_pos_allocation}):
First,  we need identify robots sharing the same current target node, i.e., \(\operatorname{Node}^{(n)}[1]\), and group them into a set \(\theta\), where \(n \in \theta\) (line 2).  Denote the target node as \(v_{\operatorname{tar}} = \operatorname{Node}^{(n)}[1]\), with its set of optional positions denoted as \(E_{v_{\operatorname{tar}}}\) and position indices \(e \in E_{v_{\operatorname{tar}}}\). The position allocation aims to prevent excessive assignments to any single position while ensuring each robot is assigned exactly one position.

To achieve this, we introduce a weighted cost formulation that balances the distance cost and the degree of current position occupancy.
For each robot \(n \in \theta\) and position \(e \in E_{v_{\operatorname{tar}}}\), we define:
\(d_{n,e}\): the euclidean distance from the current position of robot \(n\) to position \(e\);
\(\mu_e\): the number of robots assigned to position \(e\), initialized to 0;
\(C_v\): the reference capacity of the node, computed as \(C_v = |\theta| / |E_{v_{\operatorname{tar}}}|\), representing the expected number of robots per position.
Then, the comprehensive cost function for assigning robot \(n\) to position \(e\) is:
\begin{equation}
	\operatorname{Cost}_{n,e} = d_{n,e} + \tau \frac{\mu_e}{C_v},
\end{equation}
where \(\tau\) is a tuning parameter balancing the trade-off between distance cost and position occupancy.
The allocation employs a greedy strategy: for each robot \(n \in \theta\), compute the cost \(\operatorname{Cost}_{n,e}\) for all positions \(e \in E_{v_{\operatorname{tar}}}\), select the position with the minimum cost:
\begin{equation}
	e^* = \arg \min_{e \in E_{v_{\operatorname{tar}}}} \operatorname{Cost}_{n,e},
\end{equation}
assign robot \(n\) to \(e^*\), and update \(\mu_{e^*} \gets \mu_{e^*} + 1\). This process repeats until all robots in \(\theta\) are assigned (line 5-7).

\begin{algorithm}[t]
	\caption{Position allocation}
	\label{alg:path_pos_allocation}
	\begin{algorithmic}[1]
		\State \textbf{Input:} $\operatorname{Node}^{(n)}$
		\State Identify all robot group $\theta$.
		\State Initialize $\mu_e \gets 0$ for each $e \in E_{v_{\operatorname{tar}}}$, $C_v \gets |\theta| / |E_{v_{\operatorname{tar}}}|$
		\For{each $n \in \theta$}
		\State Find $e^* \gets \arg\min_{e \in E_{v_{\operatorname{tar}}}} \left( d_{n,e} + \tau \cdot \frac{\mu_e}{C_v} \right)$
		\State Assign $n$ to $e^*$, update $\mu_{e^*} \gets \mu_{e^*} + 1$
		\State  $\operatorname{Pos}^{(n)}[1] \gets e^*$
		\For{each $v_i = \operatorname{Node}^{(n)}[i]$, $i \geq 2$}
		\State $\operatorname{Pos}^{(n)}[i] \gets$
		\Statex \quad \quad \quad $\arg\min_{e \in E_{v_i}} \text{Distance}(\operatorname{Pos}^{(n)}[i-1], e)$
		\EndFor
		\EndFor
		\State \textbf{Output:} $\operatorname{Pos}^{(n)}$
	\end{algorithmic}
\end{algorithm}

For subsequent nodes \(v[i] = \operatorname{Node}^{(n)}[i]\) (\(i \geq 2\)), a simplified greedy strategy is adopted: select the position from \(E_{v_i}\) that minimizes the euclidean distance to the position assigned to the previous node, \(\operatorname{Pos}^{(n)}[i-1]\) (line 8-10). This approach prevents overcrowding at the first node through the comprehensive cost function, while the distance-based greedy selection for subsequent nodes significantly reduces computational complexity, ensuring allocation efficiency.
Through this method, we successfully assign the corresponding \(\operatorname{Pos}^{(n)}\) to \(\operatorname{Node}^{(n)}\), generating the final discrete path.

For the initial node allocation, a weighting factor was applied, while subsequent allocations used a nearest-distance strategy. This approach ensures minimal computational cost for online execution and avoids over-complicated decisions for distant goals, whose environments may change significantly before arrival.

\section{Experiments}

In this section, we demonstrate the effectiveness of the proposed method. We design two groups of experiments under different scenarios to evaluate its performance. All experiments run on a desktop computer equipped with an Intel Core i7-12700 CPU and 32 GB of RAM.

We divide the experiments into small-scale scenarios with 10–100 robots and large-scale scenarios with 100–500 robots. Since different methods have distinct limitations in applicability and scalability, we select different sets of baseline algorithms for each scale to ensure a comprehensive comparison.
Table~\ref{tab:method-comparison} summarizes all algorithms considered in the comparison.
In both groups, we use A* as a baseline. In this setting, each robot independently computes its shortest path, serving as a reference for individually optimal path planning. Although simple, this baseline highlights the challenges of resolving conflicts in multi-robot systems. The point-to-point planning algorithm is excluded due to its consistently poor performance, which provides little meaningful evaluation.
For small-scale scenarios, we include ECBS and P-SIPP as representative MAPF algorithms. These classic benchmarks are widely used in the field but their success rate drops sharply when the number of robots exceeds 100, so we only consider them in small-scale experiments. 
TVMA, another congestion-aware baseline, is only included in the small-scale experiments because it exhibits poor scalability; its computational cost grows rapidly with the number of robots, and the original paper reports results with no more than 15 robots.
For large-scale scenarios, we select CBSH2-RTC and LNS2, which represent the state-of-the-art in the MAPF field. These methods solve large-scale planning problems in complex environments, showing better scalability and higher solution quality. 
We also include CMPP, another congestion-aware method, in the large-scale experiments to further benchmark the proposed approach.

\begin{table}[t]
	\centering
	\caption{Comparison with front-end discrete path planning methods.}
	\label{tab:method-comparison}
	\resizebox{\linewidth}{!}{
		\begin{tabular}{l *{4}{c}}
			\toprule
			Method & Multi-agent & Online & Large-scale & Congestion \\
			\midrule
			Point-to-Point & $\times$ & $\times$ & $\checkmark$ & $\times$ \\
			A* & $\times$ & $\times$ & $\checkmark$ & $\times$ \\
			ECBS \cite{ECBS} & $\checkmark$ & $\times$ & $\times$ & $\times$ \\
			P-SIPP \cite{p-sipp} & $\checkmark$ & $\times$ & $\times$ & $\times$ \\
			CBSH2-RTC \cite{CBSH2-RTC} & $\checkmark$ & $\times$ & $\checkmark$ & $\times$ \\
			LNS2 \cite{LNS2} & $\checkmark$ & $\times$ & $\checkmark$ & $\times$ \\
			TVMA \cite{TVMA} & $\checkmark$ & $\times$ & $\times$ & $\checkmark$ \\
			CMPP \cite{CMPP} & $\checkmark$& $\times$ & $\checkmark$ & $\checkmark$ \\
			\textbf{Flow (Ours)} & $\checkmark$ & $\checkmark$ & $\checkmark$ & $\checkmark$ \\
			\bottomrule
	\end{tabular}}
	\captionsetup{font=large} 
	\caption*{\small Note: Multi-agent indicates whether the algorithm considers coordination among multiple robots. Online indicates support for online planning. Large-scale indicates capability to handle scenarios with over 100 robots. Congestion indicates whether the algorithm supports trade-offs between queuing and detouring to alleviate congestion.}
\end{table}

\begin{figure}[t]
	\centering
	\begin{subfigure}[b]{0.23\textwidth}
		\includegraphics[width=\textwidth]{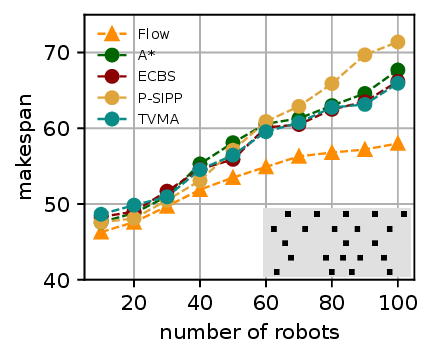}
		\caption{Forest}
		\label{fig-num-time-f-1}
	\end{subfigure}
	\begin{subfigure}[b]{0.23\textwidth}
		\includegraphics[width=\textwidth]{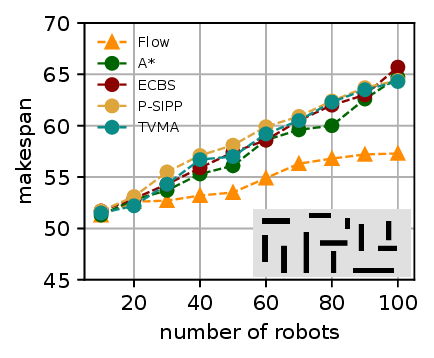}
		\caption{Maze}
		\label{fig-num-time-m-1}
	\end{subfigure}
	\begin{subfigure}[b]{0.23\textwidth}
		\includegraphics[width=\textwidth]{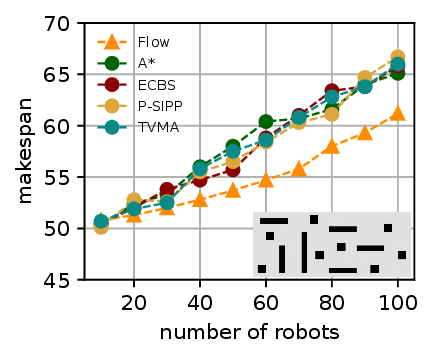}
		\caption{Clutter}
		\label{fig-num-time-c-1}
	\end{subfigure}
	\begin{subfigure}[b]{0.23\textwidth}
		\includegraphics[width=\textwidth]{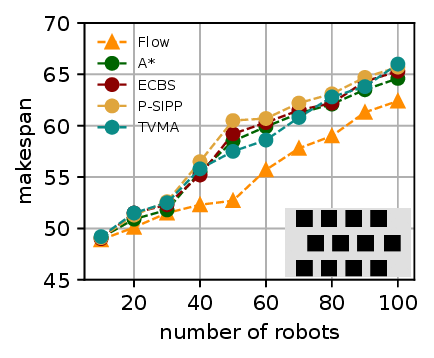}
		\caption{Corridor}
		\label{fig-num-time-w-1}
	\end{subfigure}
	\caption{Makespan results w.r.t. number of robots. Figures (a–d) show the makespan results for four different maps with 10 to 100 robots. An example of the corresponding map type is shown in the lower right corner of each image.}
	\label{fig-num-time-small}
\end{figure}

\begin{table}[t]
	\centering
	\caption{Average improvement of the Flow in the range 10--100 (unit: \%).}
	\label{tab:flow-improvement-10-100}
	\small
	\sisetup{output-decimal-marker={.}, group-digits=false}
	\resizebox{0.7\linewidth}{!}
	{
		\begin{tabular}{@{} l *{4}{S[table-format=2.2]} @{}}
			\toprule
			Map & {A*} & {ECBS} & {P-SIPP} & {TVMA} \\
			\midrule
			Forest   & 7.92 & 7.10 & 9.36 & 7.05 \\
			Maze     & 5.02 & 6.21 & 6.98 & 6.14 \\
			Clutter  & 5.55 & 5.15 & 5.21 & 5.33 \\
			Corridor & 5.37 & 5.97 & 6.87 & 4.67 \\
			\bottomrule
	\end{tabular}}
\end{table}

\begin{table}[t]
	\centering
	\small
	\caption{Computation time for small-scale scenarios under different node sizes and robot teams (10–100) (unit: seconds).}
	\label{tab:runtime-small-flow}
	\resizebox{0.8\linewidth}{!}{
		\begin{tabular}{cc 		S[table-format=4.1] 
				S[table-format=4.1]  
				S[table-format=2.0]  
				S[table-figures-integer=1,table-figures-decimal=2,table-number-alignment=left]} 
			\toprule
			\multirow{2}{*}{\text{Nodes $\left| \mathcal{V} \right|$}} & \multirow{2}{*}{\text{Robots $N$}}
			& \text{ECBS} & \text{P-SIPP} & \text{TVMA} & \text{Flow} \\
			\cmidrule(lr){3-6}
			\multirow{3}{*}{50}
			& 10  & 1 & 2 & 6 & 0.04 \\
			& 50  & 3 & 5 & 22 & 0.10 \\
			& 100 & 6 & 10 & 180 & 0.13\\
			\midrule
			\multirow{3}{*}{100}
			& 10  & 2 & 2 & 15 & 0.06 \\
			& 50  & 5 & 5 & 61 & 0.12 \\
			& 100 & 8 & 16 & 280 & 0.14 \\
			\midrule
			\multirow{3}{*}{150}
			& 10  & 3 & 4 & 24 & 0.08 \\
			& 50  & 7 & 15 & 67 & 0.14 \\
			& 100 & 15 & 20 & 312 & 0.16 \\
			\bottomrule
	\end{tabular}}
\end{table}

\subsection{Small-scale scenarios (10–100 robots)}

In the first set of experiments, we vary the number of robots from 10 to 100 (i.e., 10, 20, 30, 40, 50, 60, 70, 80, 90, 100). For each robot count, we test the algorithms on four different map types: forest, maze, clutter, and corridor. 
For each type of map, five map instances are generated  respectively.
The algorithms evaluated in this set are:

\noindent (a) A*,
(b) ECBS\cite{ECBS}, 
(c) P-SIPP \cite{p-sipp},
(d) TVMA \cite{TVMA}.

The experimental parameters are configured as follows. The minimum safety distance is set to $r_{\min} = 0.4\text{m}$. In the path set search module, the parameters are chosen as $\alpha = 5$ and $\beta = 5$. The time discretization step is $h = 0.1s$, and the maximum velocity is limited to $\mathbf{v}_{\max} = 5\text{m/s}$.
In the network construction, the parameters are set as $N_B = 4$, $L_{\text{con}} = 5$, $W_{\text{con}} = 5$, and $\phi = 1.5$.
Regarding the predictive path length and sub-segment settings, we use $L_{\text{pre}} = 15$, $K = 3$. (The parameter settings for $L_{\text{pre}}$ and $K$ are detailed in Fig.  \ref{fig-makespace-K}.)
The weight coefficients are set to:  $\omega_1=1$, $\omega_2=1$, $\omega_3=1$,  $\omega_{\text {run }}=0.01$ and $\tau = 0.5$.
The initial and goal positions of the robots are uniformly arranged in a matrix queue format.
Our collision avoidance algorithm in the back-end module utilizes the ORCA \cite{RVO} method, chosen for two primary reasons. Firstly, ORCA demonstrates reliable performance in multi-robot collision avoidance, reducing the likelihood of deadlocks. Secondly, it offers high computational efficiency, maintaining acceptable computation times even with a large number of robots.
The frequency of collision avoidance algorithm is 10 Hz.
Our flow planner is triggered periodically every 2 seconds.

Fig. \ref{fig-num-time-small} illustrates the makespan performance of our algorithm under small-scale robot numbers across different map types, with Table~\ref{tab:flow-improvement-10-100} reporting their average improvement ratios.
Based on the results, we make the following observations:
(1) In experiments with 10–100 robots, our Flow algorithm achieves the most efficient makespan performance, while P-SIPP yields the poorest results. ECBS outperforms A*, and TVMA performs better than ECBS.
(2) This demonstrates that Flow handles congestion caused by increasing robot density more effectively.
(3) When the number of robots is small, congestion is minimal, and the makespan performance of all four methods is comparable.
(4) As $N$ increases, the overall time cost rises, and the performance gap between Flow and other methods widens. This is because higher robot density leads to more severe congestion and longer waiting times. By replanning for waiting robots, Flow reduces overall traversal time, making its advantage increasingly pronounced as $N$ grows.
(5) Traditional MAPF algorithms generally perform poorly as front-end planners. Their generated paths often consist of right-angle turns, which become inefficient for long-distance navigation across maps.

Table~\ref{tab:runtime-small-flow} reports the computation time under different map sizes (number of nodes) and varying numbers of robots. For ECBS, P-SIPP, and TVMA, which are offline methods, the reported computation time corresponds to the total runtime of a single offline planning process. In contrast, Flow is an online method. Its computation time is measured by recording the duration of each planning step, summing them, and dividing by the total number of planning steps to obtain the average planning time.
The results reveal a clear difference in computation efficiency. Flow demonstrates high efficiency in small-scale scenarios, with average planning times typically below 0.2 seconds, enabling rapid decision-making. By contrast, TVMA’s computation time grows quickly with the number of robots, resulting in poor scalability.

\subsection{Large-scale scenarios (100–500 robots)}

The second set of experiments is designed to evaluate algorithm performance under large-scale conditions. The number of robots is increased from 100 to 500 in increments of 50 (i.e., 100, 150, 200, 250, 300, 350, 400, 450, 500). Similar to the first set, we test on the same four types of maps. The algorithms compared in this set are:

\noindent (a) A*,
(b) CBSH2-RTC \cite{CBSH2-RTC},
(c) LNS2 \cite{LNS2},
(d) CMPP \cite{CMPP}.

\begin{figure}[t]
	\centering
	\begin{subfigure}[b]{0.23\textwidth}
	\includegraphics[width=\textwidth]{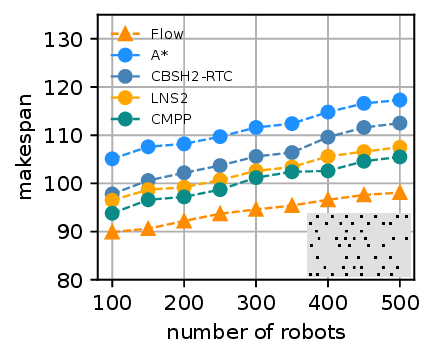}
	\caption{Forest}
	\label{fig-num-time-f-2}
	\end{subfigure}
	\begin{subfigure}[b]{0.23\textwidth}
		\includegraphics[width=\textwidth]{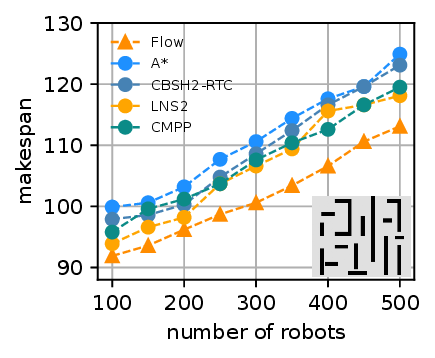}
		\caption{Maze}
		\label{fig-num-time-m-2}
	\end{subfigure}
	\begin{subfigure}[b]{0.23\textwidth}
		\includegraphics[width=\textwidth]{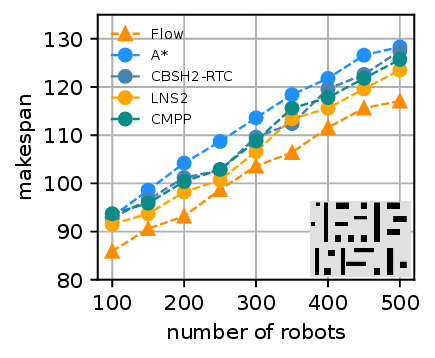}
		\caption{Clutter}
		\label{fig-num-time-c-2}
	\end{subfigure}
	\begin{subfigure}[b]{0.23\textwidth}
		\includegraphics[width=\textwidth]{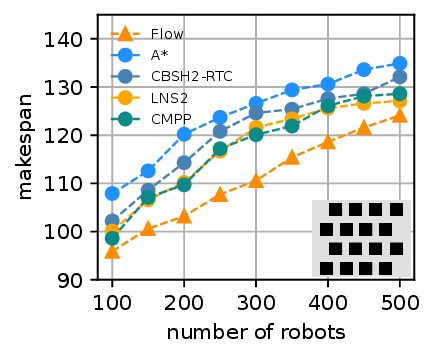}
		\caption{Corridor}
		\label{fig-num-time-w-2}
	\end{subfigure}
	\caption{Makespan results w.r.t. number of robots. Figures (a–d) show the makespan results for four different maps with 100 to 500 robots.}
	\label{fig-num-time-large}
\end{figure}

\begin{table}[t]
	\centering
	\caption{Average improvement of the Flow in the range 100--500 (unit: \%).}
	\label{tab:flow-improvement-100-500}
	\sisetup{output-decimal-marker={.}, group-digits=false}
	\resizebox{0.7\linewidth}{!}{
		\begin{tabular}{@{} l *{4}{S[table-format=2.2]} @{}}
			\toprule
			Map & {A*} & {CBSH2-RTC} & {LNS2} & {CMPP} \\
			\midrule
			Forest   & 15.40 & 7.83 & 6.66 & 5.97 \\
			Maze     & 8.39 & 6.84 & 4.58 & 5.48 \\
			Clutter  & 8.94 & 4.16 & 6.32 & 6.09 \\
			Corridor & 9.87 & 7.97 & 5.69 & 5.64 \\
			\bottomrule
	\end{tabular}}
\end{table}

\begin{table}[t]
	\centering
	\small
	\caption{Computation time for large-scale scenarios under different node sizes and robot teams (100–500) (unit: seconds).}
	\label{tab:runtime-large-flow}
	\resizebox{0.9\linewidth}{!}{
		\begin{tabular}{cc 
				S[table-format=4.1]  
				S[table-format=4.1]  
				S[table-format=2.0]  
				S[table-figures-integer=1,table-figures-decimal=2,table-number-alignment=left]} 
			\toprule
			\multirow{2}{*}{\text{Nodes $\left| \mathcal{V} \right|$}} & \multirow{2}{*}{\text{Robots $N$}}
			& \text{CBSH2-RTC} & \text{LNS2} & \text{CMPP} & \text{Flow} \\
			\cmidrule(lr){3-6}
			\multirow{3}{*}{100}
			& 100 & 3 & 2 & 60 & 0.11 \\
			& 300 & 11 & 8 & 60 & 0.26 \\
			& 500 & 37 & 55 & 60 & 0.38 \\
			\midrule
			\multirow{3}{*}{300}
			& 100 & 3 & 2 & 60 & 0.17 \\
			& 300 & 16 & 13 & 60 & 0.45 \\
			& 500 & 42 & 58 & 60 & 0.79 \\
			\midrule
			\multirow{3}{*}{500}
			& 100 & 4 & 2 & 60 & 0.23 \\
			& 300 & 21 & 15 & 60 & 0.51 \\
			& 500 & 58 & 62 & 60 & 0.91 \\
			\bottomrule
	\end{tabular}}
\end{table}

Fig. \ref{fig-num-time-large} illustrates the makespan performance of our algorithm across different map types with varying numbers of robots, with Table~\ref{tab:flow-improvement-100-500} reporting their average improvement ratios.
From the figures, we observe the following:
(1) In the experiments with 100–500 robots, Flow consistently achieves the best makespan performance.  A* performs the worst, followed by CBSH2-RTC and LNS2, CMPP while  performs relatively better.
(2) As an advanced MAPF method, LNS2 alleviates congestion through optimized front-end path planning, thereby improving the efficiency of the control module in the subsequent execution phase. In contrast, CBSH2-RTC is less effective in this regard.
(3) The CMPP method is effective in reducing congestion, but since its cost function only considers congestion, some robots tend to take excessive detours, which reduces overall traversal efficiency.
(4) As the number of robots increases from 100 to 500, the performance gap between Flow and the other algorithms gradually widens, indicating that Flow can continue to effectively handle congestion in large-scale scenarios.

Table.~\ref{tab:runtime-large-flow} reports the computation time under varying map sizes (number of nodes) and robot populations.
For CBSH2-RTC, LNS2, and CMPP, which are offline methods, the reported values represent the total runtime of a complete offline planning process. In contrast, the values for Flow correspond to the average planning time.
Due to its algorithmic design, CMPP continuously expands nodes to obtain improved solutions. As a result, longer expansion times lead to better outcomes. Following the setting in the original paper, we fix the expansion time to 60 seconds in our experiments.
The results show that Flow requires only 0.38 seconds when the map size is 100 nodes with 500 robots. Even with 500 nodes, the computation time remains below 1 second. These findings demonstrate that, within a reasonable node limit, the Flow algorithm is capable of meeting real-time requirements.

\begin{table}[t]
	\centering
	\large
	\caption{Component-wise computation time of Flow (unit: seconds).}
	\label{tab:computation-time}
	\resizebox{\linewidth}{!}{
		\begin{tabular}{l *{7}{S[table-format=1.3]}}
			\toprule
			& \multicolumn{7}{c}{The number of robots} \\
			\cmidrule(lr){2-8}
			Method & {10} & {50} & {100} & {200} & {300} & {400} & {500} \\
			\midrule
			Path set search & 0.048 & 0.061 & 0.101 & 0.161 & 0.251 & 0.321 & 0.416 \\
			Path node selection & 0.015 & 0.060 & 0.121 & 0.180 & 0.246 & 0.327 & 0.467 \\
			Position allocation & 0.000 & 0.002 & 0.004 & 0.008 & 0.015 & 0.013 & 0.025 \\
			\midrule
			\textbf{Total} & 0.063 & 0.123 & 0.226 & 0.349 & 0.512 & 0.661 & 0.908 \\
			\bottomrule
	\end{tabular}}
\end{table}

\sisetup{table-number-alignment=center, group-separator={,}}

\begin{figure}[t]
	\centering
	\includegraphics[scale=0.6]{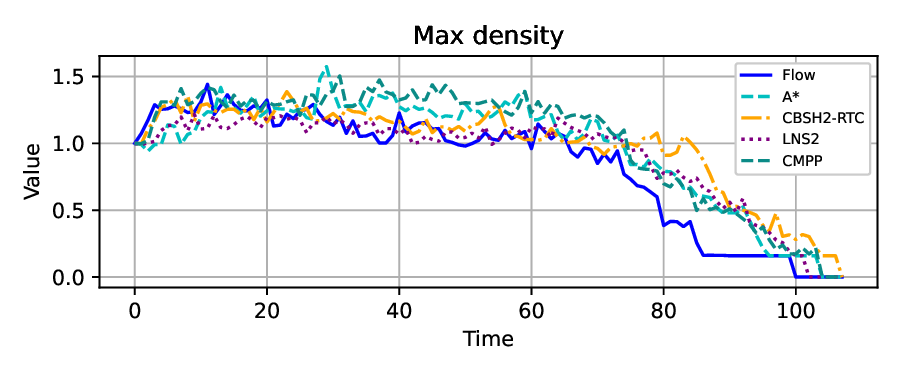}
	\includegraphics[scale=0.6]{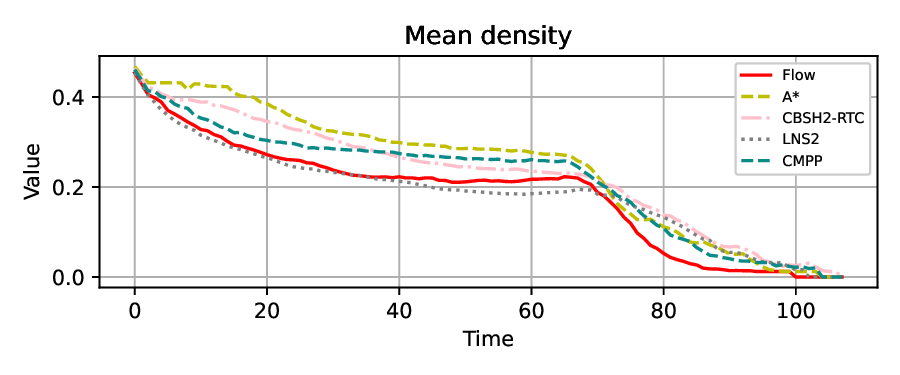}
	\caption{Maximum and mean robot densities over time in the forest map with 500 robots.}
	\label{fig:density_comparison}
\end{figure}
\begin{figure}[t]
	\centering
	\includegraphics[scale=0.6]{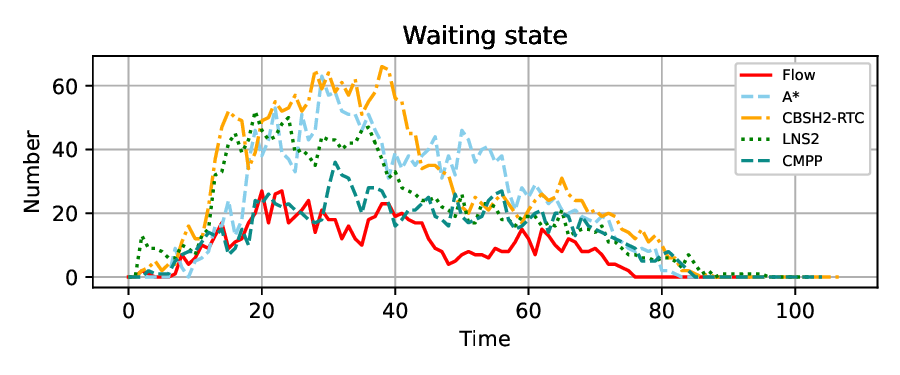}
	\caption{The number of robots in the waiting state over time under the forest map with 500 Robots.}
	\label{fig:wait robots}
\end{figure}

\begin{figure}[t]
	\centering
	\begin{subfigure}[b]{0.23\textwidth}
		\includegraphics[width=\textwidth]{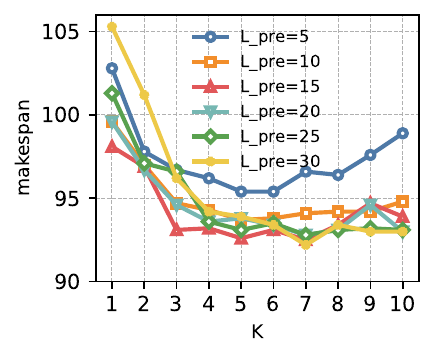}	
	\end{subfigure}
	\begin{subfigure}[b]{0.23\textwidth}
		\includegraphics[width=\textwidth]{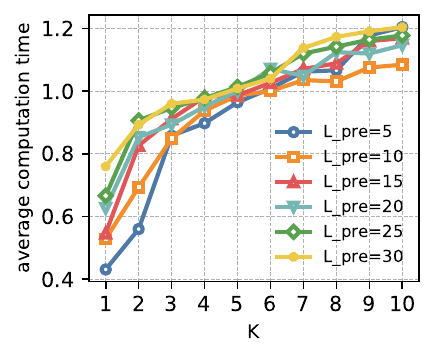}
	\end{subfigure}
	\caption{Makespan results and average computation time  w.r.t. different values of $K$ and $L_{\text{pre}}$ on the forest map of 500 robots.}
	\label{fig-makespace-K}
\end{figure}

Table \ref{tab:computation-time} reports the component-wise computation time of the Flow algorithm. 
The results reveal the following:
(1) The majority of computation time is spent on path set search and path node selection, whereas position allocation incurs minimal overhead. This is consistent with the algorithm's design, where the final position allocation step is deliberately kept simple and efficient to minimize computational cost.
(2) When the number of robots $N$ is 10, the total computation time is approximately 0.06 seconds; for $N = 500$, it increases modestly to around 0.9 seconds. These results demonstrate that the Flow algorithm achieves high computational efficiency, making it well-suited for real-time online planning scenarios.

We further analyze the spatial distribution characteristics and temporal evolution of robot density under different algorithms in a forest environment with a total of 500 robots.
The evaluation procedure is as follows: the simulated environment is discretized into multiple grid cells, and the number of robots within each cell is accurately recorded to construct the raw density distribution map. Based on this map, we compute two key metrics for each region: the maximum density and the non-zero mean density. The corresponding results are shown in Fig.  \ref{fig:density_comparison}.
Our observations are summarized as follows:
(1) At the early stage of the simulation, the differences in maximum density among the five algorithms are relatively small. However, after 30 time units, these differences become increasingly significant.
Specifically, the maximum density under the A* algorithm rises rapidly, reaching the highest level among all methods, followed by CBSH2-RTC and CMPP.
In contrast, LNS2 and the proposed Flow algorithm maintain relatively lower peak densities.
This trend indicates that A*, due to its strict enforcement of shortest-path planning for each robot, tends to induce severe congestion in the early stage.
Meanwhile, LNS2 demonstrates a certain capability in alleviating congestion, performing comparably to the Flow algorithm during the initial phase.
(2) By the 60th time unit, the maximum density under the Flow algorithm shows a marked and rapid decrease. This result highlights the effectiveness of the Flow algorithm’s early-stage planning strategy, which significantly reduces peak density in later stages, thereby promoting a more balanced spatial distribution of robots across the environment.
(3) Regarding non-zero mean density, LNS2 and the Flow algorithm yield comparable values before the 70th time unit, with LNS2 achieving slightly lower averages. However, LNS2 displays a clear drawback in makespan efficiency. This indicates that although LNS2 can alleviate congestion to some extent, its solutions often involve considerable detours, leading to longer travel times. In contrast, the Flow algorithm effectively balances congestion avoidance with path length, delivering superior overall performance in terms of time efficiency.

To further evaluate system performance, we define a robot as being in a waiting state if its speed falls below $0.5\mathrm{m/s}$. Based on this criterion, we analyze the temporal evolution of the number of robots in the waiting state within the forest scenario of 500 robots. The corresponding results are presented in Fig.  \ref{fig:wait robots}. Statistical analysis reveals that the Flow algorithm consistently maintains the lowest number of waiting robots among all five methods.
Combined with the earlier density analysis, these findings demonstrate that the Flow algorithm effectively balances path congestion and robot waiting time. It enables efficient utilization of road resources while minimizing extended idle periods. Notably, the Flow algorithm does not rely on simplistic detour strategies; rather, it achieves high-throughput navigation in cluttered environments by maximizing roadway capacity utilization and ensuring smooth robot traversal in obstacle-rich regions.

To analyze the impact of different values of the algorithmic parameters $L_{\text{pre}}$ and $K$, we conduct experiments on a forest map scenario involving 500 robots. We record the resulting makespan and average computation time under various parameter configurations. The results are presented in Fig. \ref{fig-makespace-K}. Based on these results, we draw the following observations and recommendations for selecting suitable parameter values:
(1) When $L_{\text{pre}} = 5$, the makespan is significantly longer. This suggests that a predicted initial trajectory that is too short limits the algorithm’s ability to achieve satisfactory performance.
(2) When $L_{\text{pre}}$ exceeds 10, the makespan is noticeably reduced. In particular, increasing $K$ from 1 to 3 leads to a substantial improvement in performance. However, further increasing $K$ yields diminishing returns in terms of makespan reduction.
(3) The average computation time increases significantly with larger values of $K$. Similarly, increasing $L_{\text{pre}}$ introduces additional computational overhead. These results indicate that higher parameter values do not necessarily translate to better overall outcomes.
(4) When $L_{\text{pre}}$ is large, using a small value of $K$ can also lead to a longer makespan. This is because a longer predicted trajectory divided into too few segments provides insufficient resolution for detecting and resolving congestion.
Based on these findings, we conclude that a balanced choice of $L_{\text{pre}}$ and $K$ is essential for achieving good algorithmic performance while maintaining computational efficiency. Our experiments suggest that setting $L_{\text{pre}} = 15$ and $K = 3$ strikes an effective balance. This configuration offers significant improvement in makespan compared to smaller values of $K$, while avoiding the excessive computational costs associated with further increases.

\subsection{Real-World Flight Test}

\begin{figure}[h]
	\centering
	\includegraphics[scale=0.32]{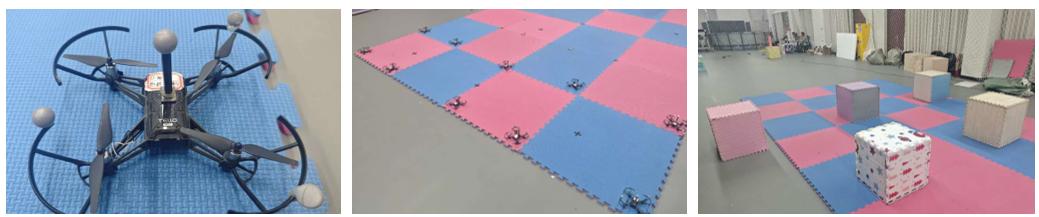}
	\caption{Real flight test scene.}
	\label{fig:expe1}
\end{figure}

\begin{figure*}[t]
	\centering
	\includegraphics[scale=0.92]{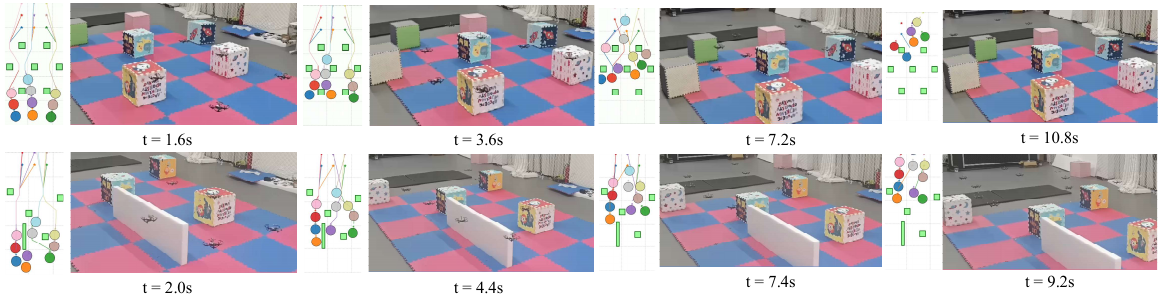}
	\caption{The flight process of swarm drones. In each pair of subfigures, the left is the simulated flight image, and the right is the corresponding real-world flight image. Each row corresponds to a specific scenario.}
	\label{fig:expe2}
\end{figure*}
We conduct flight tests using ten DJI Tello drones across three distinct scenarios, within a confined space of $15 \text{m} \times 5 \text{m} \times 2 \text{m}$ to prevent drones from bypassing obstacle regions. The real-world flight test environment is shown in Fig. \ref{fig:expe1}. A high-precision OptiTrack motion capture system provides accurate localization for all drones.
In each scenario, multiple cubic boxes (each with an edge length of $0.5\,\text{m}$) are placed as obstacles. Drones are initially positioned on one side of the scenario and navigate through the obstacle-filled region to reach the opposite side. Our proposed Flow algorithm computes real-time paths for all drones, with all computations executed on a single laptop.
The experimental parameters used in the real-world tests are as follows: $r_{\min} = 0.5\,\text{m}$, $h = 0.1\,\text{s}$, $\mathbf{v}_{\max} = 0.5\,\text{m/s}$, $N_B = 1$, $\alpha = 3$, $\beta = 3$, $L_{\text{con}} = 2$, $W_{\text{con}} = 2$, $\phi = 1.0$, $L_{\text{pre}} = 5$, $K = 2$, and $\tau = 0.5$. The weights are set as $\omega_1 = 1$, $\omega_2 = 1$, and $\omega_{\text{run}} = 1$. The collision avoidance algorithm operates at $10\,\text{Hz}$, while the scheduling algorithm runs at $1\,\text{Hz}$.
Fig. \ref{fig:expe2} illustrates the experimental scenarios, with each row representing a distinct case. For each scenario, the simulation (left) and the corresponding real-world setup (right) are displayed chronologically from left to right. Throughout all experiments, no collisions or deadlocks were observed. The detailed process of the flight tests can be seen in the supplementary video.
These experiments across all three scenarios validate the effectiveness and practical deployability of our algorithm on real-world robotic platforms.

\section{Conclusion}

In this paper, we proposed a flow-inspired scheduling planner to optimize the traversal of multi-robot systems in obstacle-rich environments. By exploiting the traversable areas of the environment and organizing robot motions in a flow-like manner, the proposed planner improves traversal efficiency while maintaining high computational efficiency suitable for real-time deployment. Extensive simulations and real-world drone experiments demonstrate the effectiveness and practicality of the proposed framework.
It is worth noting that the current approach adopts a relatively sparse network construction that emphasizes unidirectional robot flows. To facilitate efficient flow-based scheduling, the modeling does not explicitly capture the path diversity required for omnidirectional robot movements. As a result, the proposed method is particularly well suited for traversal tasks in which robots move from one side of the environment to the other. Extending the flow-inspired scheduling framework to support omnidirectional and multi-directional robot flows remains an important direction for future research.

\bibliographystyle{unsrt}
\bibliography{citepaper_all}

\end{document}